\documentclass[journal]{IEEEtran}
\usepackage{amsmath,amsfonts}
\usepackage{algorithmic}
\usepackage{algorithm}
\usepackage{array}
\usepackage{textcomp}
\usepackage{stfloats}
\usepackage{url}
\usepackage{verbatim}
\usepackage{ulem}
\usepackage{graphicx}
\usepackage{cite}
\usepackage[caption=false,font=footnotesize,labelfont=rm,textfont=rm]{subfig}
\usepackage{color}
\usepackage{lipsum}
\usepackage{multirow}
\usepackage{booktabs}
\usepackage{stfloats}
\usepackage{amssymb}
\DeclareMathOperator{\st}{s.t.}
\usepackage{cleveref}

\usepackage{titlesec}
\titleformat{\paragraph}[runin]
  {\normalfont\normalsize\bfseries}{}{0pt}{}[.\hspace{0.5em}\hspace{0.5em}]
\titlespacing{\paragraph}{0pt}{0pt}{0pt}

\renewcommand{\uwave}{}
\begin{document}
\title{HeightFormer: Explicit Height Modeling \\
without Extra Data for Camera-only \\
3D Object Detection in Bird’s Eye View}

\author{
  
       Yiming~Wu$^\dag$, Ruixiang~Li$^\dag$, Zequn~Qin$^{*}$, Xinhai~Zhao, Xi~Li$^{*}$

\IEEEcompsocitemizethanks{\IEEEcompsocthanksitem
Yiming~~Wu is with Polytechnic Institute, Zhejiang University, Hangzhou 310015, China.
Ruixiang~Li, Zequn~Qin, and Xi~Li are with College of Computer Science and Technology, Zhejiang University, Hangzhou 310027, China.
Xinhai~Zhao is with Noah's Ark Lab, Huawei Technologies, Shanghai 201206, China.
E-mail: \{nolva,ruixli\}@zju.edu.cn, zequnqin@gmail.com, zhaoxinhai1@huawei.com, xilizju@zju.edu.cn.
\protect
}

\thanks{($\dag$: Equal contributions. Corresponding authors: Zequn~Qin and Xi~Li.)}}

\maketitle


\begin{abstract}
    Vision-based Bird's Eye View (BEV) representation is an emerging perception formulation for autonomous driving. The core challenge is to construct BEV space with multi-camera features, which is a one-to-many ill-posed problem. Diving into all previous BEV representation generation methods, we found that most of them fall into two types: modeling depths in image views or modeling heights in the BEV space, mostly in an implicit way. In this work, we propose to explicitly model heights in the BEV space, which needs no extra data like LiDAR and can fit arbitrary camera rigs and types compared to modeling depths. Theoretically, we give proof of the equivalence between height-based methods and depth-based methods. Considering the equivalence and some advantages of modeling heights, we propose HeightFormer, which models heights and uncertainties in a self-recursive way. Without any extra data, the proposed HeightFormer could estimate heights in BEV accurately. Benchmark results show that the performance of HeightFormer achieves SOTA compared with those camera-only methods. 
  \end{abstract}


\section{Introduction}

Vision-based Bird's Eye View (BEV) representation\cite{lu2021graph, xie2023x, yang2023bevformer, bartoccioni2023lara, lin2022sparse4d, 10557672, 10473113} is an emerging perception formulation for autonomous driving. It transforms and maps the information from the image space to a unified 3D BEV space, which can be used for various perception tasks like 3D object detection and BEV map segmentation. Moreover, the unified BEV space can directly fuse other modalities like LiDAR without any cost, which is of great scalability.

\begin{figure}[!t]
    \centering
    \subfloat[The ill-posed 2D to 3D mapping problem.]{
    \includegraphics[width=0.9\linewidth]{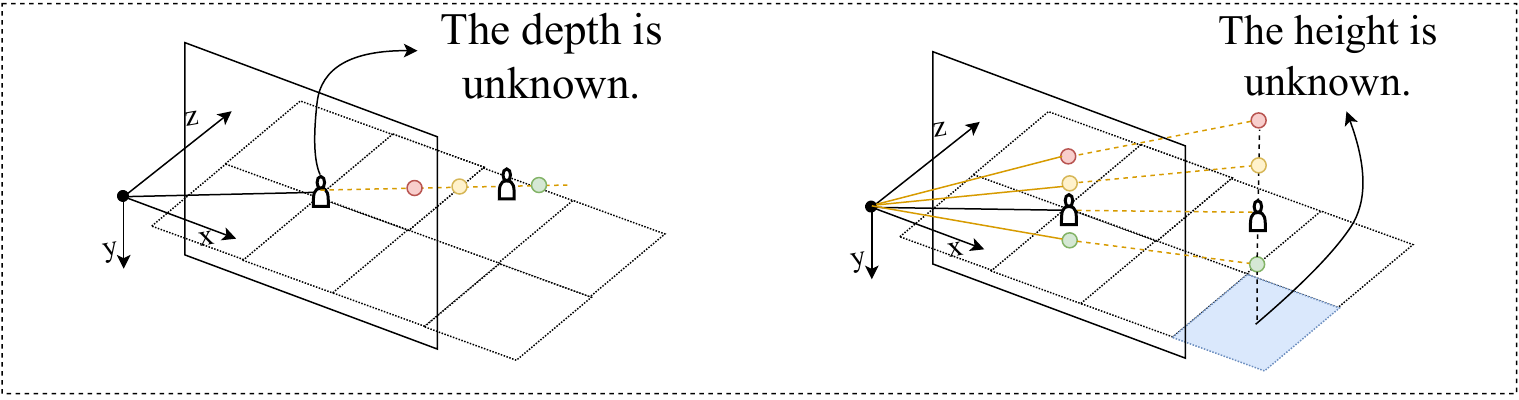}}
    
    \subfloat[Two ways of adding the extra condition.]{\includegraphics[width=0.9\linewidth]{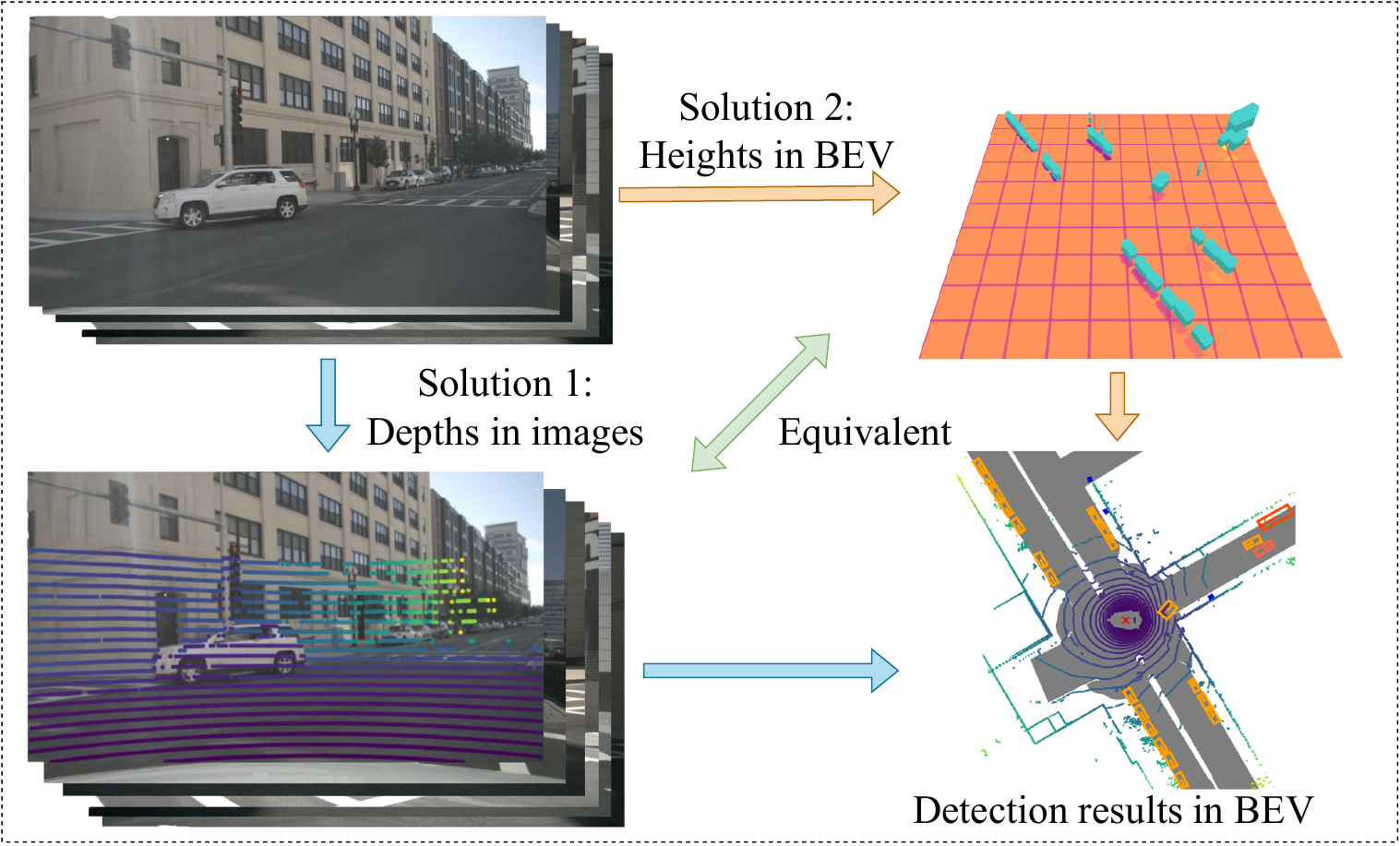}}
    \caption{Well-posed 2D to 3D mapping with the extra condition added. In solution 1, extra depth information of images is introduced. In solution 2, extra height information in BEV is introduced.
    }
    \label{fig:first}
\end{figure}

As shown in \cref{fig:first}, the essence of BEV representation is the 2D to 3D mapping, which is a one-to-many ill-posed problem because a point in the image space corresponds to infinite collinear 3D points along the camera ray. To resolve this problem, we need to add an extra condition to make the 2D to 3D mapping a one-to-one well-posed problem. For the added extra condition, there are two kinds of methods, which are LSS\cite{philion2020lift} and OFT\cite{roddick2018orthographic}. LSS proposes to predict latent depth as the extra condition, which is implicitly estimated by end-to-end training. OFT directly maps the 2D information to 3D in the one-to-many fashion, while a network in BEV space is needed to implicitly select the dense mapped information in the vertical or height direction, which is also realized by end-to-end training. Both methods use extra depth or height conditions to resolve the mapping problem, but the extra condition is implicitly trained and used. In this way, the correctness of the mapping is not guaranteed, which might affect the performance of BEV representation.

Motivated by the above observations, we propose to explicitly add and model extra conditions to realize better 2D to 3D mapping.
Similarly, some works\cite{park2021pseudo,li2022bevdepth} propose to directly learn depth as the extra condition with depth pre-training or LiDAR information. Different from using depth, we explicitly model the height condition in the mapping for the following reasons. First, we prove that height in the BEV space is equivalent to depth in the image space for the 2D to 3D mapping problem. Both ways can provide equivalent conditions to resolve the problem of mapping. In this way, we can realize well-defined one-to-one mapping between 2D and 3D. Second, the height information in the BEV space can be retrieved from the BEV annotations without any other data modalities like LiDAR, while depth condition needs extra pre-training or LiDAR. In this work, we use the height information from the object's 3D bounding box, which can be directly accessed from the ground truth. Third, the modeling in height can fit arbitrary camera rigs and types. For example, on NuScenes\cite{nuscenes2019}, the focal length of the backward camera is different from other cameras, resulting in different depth estimation patterns. In other words, different depth estimation network is needed for different cameras. For the height condition, no matter which kind of camera configuration is used, it is processed with the same pattern in the BEV space. In this way, the height condition is more robust and flexible.

In this work, we propose a network that explicitly models height in the BEV space, which fulfills the condition needed for 2D to 3D mapping, termed as HeightFormer. Moreover, based on the height modeling, self-recursive height predictors are proposed to introduce the uncertainty of heights and segmentation maps which are used in the BEV query mask mechanism to produce high-precision detection results. In summary, the main contributions of our work are summarized as follows:
    1) We give theoretical proof of the equivalence between height-based methods in BEV and depth-based methods in images, which is the basis of our work. The proof also demonstrates the feasibility of detection in the BEV space generated with predicted heights.
    2) {{\color{blue}}We propose to explicitly model heights in the BEV space without extra LiDAR supervision. A self-recursive predictor is proposed to model heights and a corresponding segmentation-based query mask is designed to handle positions whose heights cannot be defined.
    3) Experiments on NuScenes\cite{nuscenes2019} show that the proposed HeightFormer achieves the SOTA performance compared with models not using LiDAR. Extensive quantitative and qualitative results show that it is feasible and effective to model heights in the BEV space and construct the BEV representation with predicted heights. The generalization analysis also shows that the proposed method can be applied to different methods, as a plugin and as compensation for depth modeling. }

\section{Related work}

\paragraph{BEV-based camera-only 3D object detection} Most recent works in camera-only 3D object detection operate in 3D space. And there are two main methods of mapping 2D features into 3D space, which are pseudo-lidar methods\cite{wang2019pseudo, you2019pseudo, ma2020rethinking} and voxel-based BEV methods\cite{roddick2018orthographic, lang2019pointpillars, yin2021center}. Due to the efficient representation of voxels, BEV methods are commonly used now. 
The key challenge of BEV-based methods is to construct the BEV space with multi-view images. OFT\cite{roddick2018orthographic} utilizes orthographic transformation to map monocular 2D features into BEV space. LSS\cite{philion2020lift} predicts depth distribution in image space and lifts 2D features to 3D space by outer product. Following LSS, CaDDN\cite{reading2021categorical} and BEVDet\cite{huang2021bevdet} utilize the depth distribution to construct the BEV representation. 

Due to the success of transformer\cite{vaswani2017attention, brown2020language, liu2021swin}, mainstream works deal with detection tasks with a transformer-like pipeline. DETR3D\cite{wang2022detr3d} follows DETR\cite{carion2020end} to generate 3D reference points from queries. To simplify the feature sampling process in DETR3D, PETR\cite{liu2022petr,liu2022petrv2} proposes 3D positional embedding and adds the temporal information to 3D PE to align different frames. 
BEVDet\cite{huang2021bevdet} uses LSS\cite{philion2020lift} method in the BEV encoder and proposes specific data augmentation and scale-NMS tricks to improve the performance. BEVFormer\cite{li2022bevformer} introduces spatial cross-attention and temporal self-attention to generate spatiotemporal grid features from history BEV information and multi-view scenes. 
To fit the nature of the ego car's perspective, PolarFormer\cite{jiang2022polarformer} advocates the exploitation of the polar coordinate system. SOLOFusion\cite{park2022time} designs an efficient but strong temporal multi-view 3D detector to leverage long-term temporal information. TBP-Former\cite{fang2023tbp} proposes a temporal BEV pyramid transformer for spatial-temporal synchronization and BEV states prediction.

\paragraph{Depth estimation} Depth estimation is essential for 3D object detection. Due to the high similarity between depth estimation and height estimation, we learn from the experiences in depth estimation. Early works focus on geometry-based methods for stereo images\cite{scharstein2002taxonomy, flynn2016deepstereo}. In the monocular situation, there are two mainstream ways to estimate depth, which are direct regression by a well-designed network\cite{eigen2014depth, fu2018deep, xu2018multi,  ding2020learning} and geometry depth derived from the pinhole imaging model\cite{cai2020monocular}. 

For the first way, many works\cite{chen2020monopair, qin2022monoground, wang2022probabilistic} adopt uncertainty to get accurate depth. Following the practice of MonoPair\cite{chen2020monopair}, most works assume that the depth follows a Laplacian distribution, and they regress it with L1 loss. 
For example, MonoFlex\cite{zhang2021objects} designs an adaptive ensemble of estimators to predict the depth of paired diagonal key points. 
In multi-view task, BEVDepth\cite{li2022bevdepth} proposes that the final detection loss establishes an implicit depth supervision and utilizes an explicit depth supervision with LiDAR points to enable a trustworthy depth estimation.

For the second way, due to the geometric relationship between heights and depths, many works try to model heights to estimate depths. GUPNet\cite{lu2021geometry} models the height distribution and uses 3D height $h_{3d}$ and 2D height $h_{2d}$ to express depth, which is easier than regressing depth directly. MonoRCNN\cite{shi2021geometry} decomposes distances into physical heights and the reciprocals of projected visual heights.


{{\color{blue}}
\paragraph{Height modeling} Compared to depth modeling, constructing BEV features via height modeling is adopted by fewer works. BEVFormer\cite{li2022bevformer} and PolarFormer\cite{chen2022polar} obtain BEV queries by sampling 3D points around fixed heights and assigning to them different attention weights, which is an implicit way of height modeling. BEVHeight\cite{yang2023bevheight} proposes to model object heights by predicting heights for image pixels and then lifts image features to BEV space via geometric transformation,
\uwave{which relies on the high camera installation and does not work well in car-side situations}.

Different from previous works, we explicitly estimate heights in BEV space and focus on car-side situations, in which cameras are mounted on the car roof and occlusion is common. \uwave{Detailed comparison can be found in the Appendix.}
This work will also prove the equivalence between the height-based solution and the depth-based solution for constructing the BEV representation.}

\begin{figure*}[!t]
    \subfloat[Depth-based BEV representation generation. An object will be correctly detected if the marked image features fall into its $\epsilon-$neighbourhood in BEV.]{\includegraphics[width=0.45\linewidth,trim=200 50 300 100,clip]{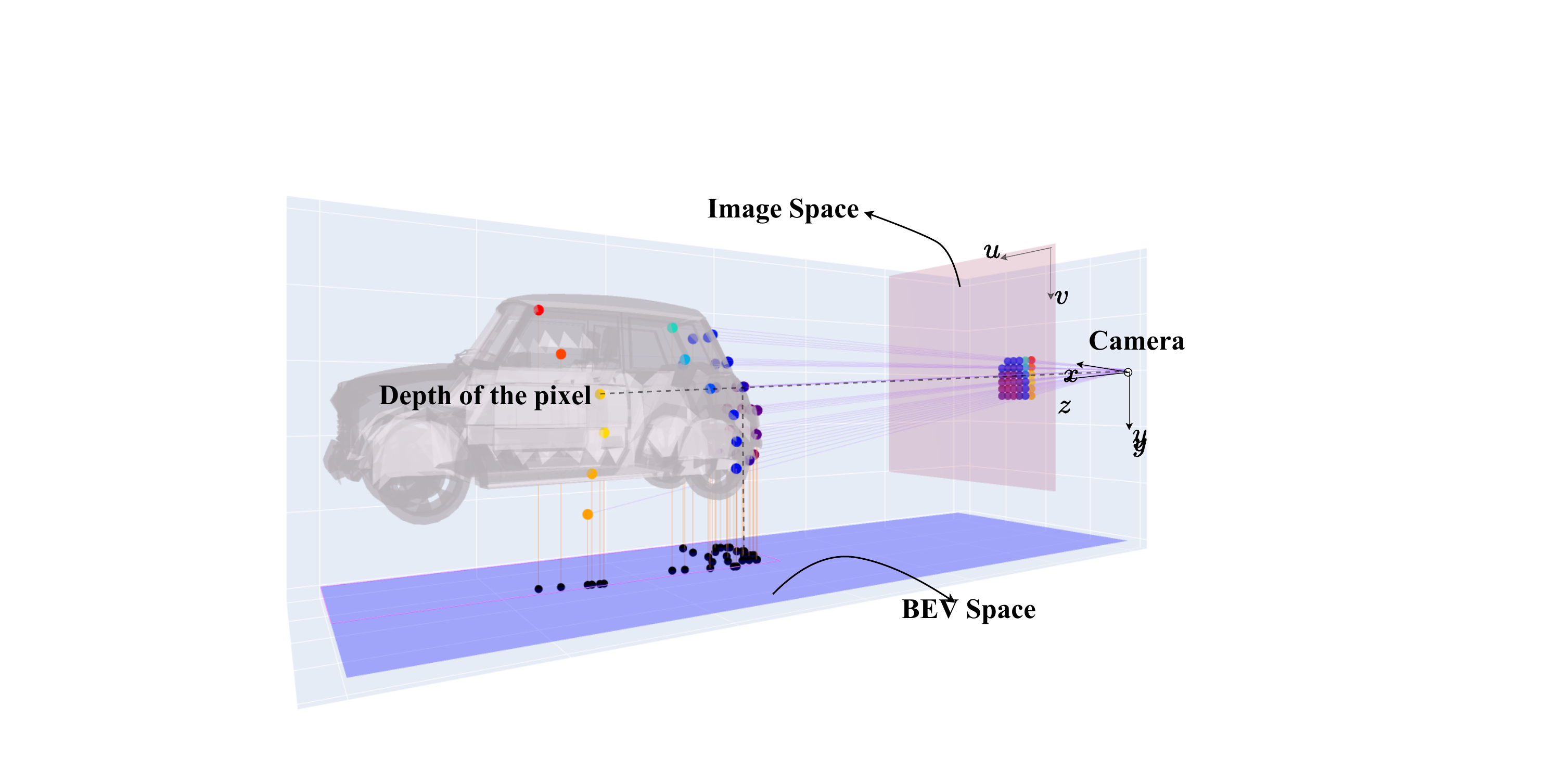}}
    \hfill
    \subfloat[Height-based BEV representation generation. An object will be correctly detected if sampled features of the $\epsilon-$neighbourhood in BEV cover the object in images. Blue lines indicate this process.]{\includegraphics[width=0.45\linewidth,trim=200 50 300 100,clip]{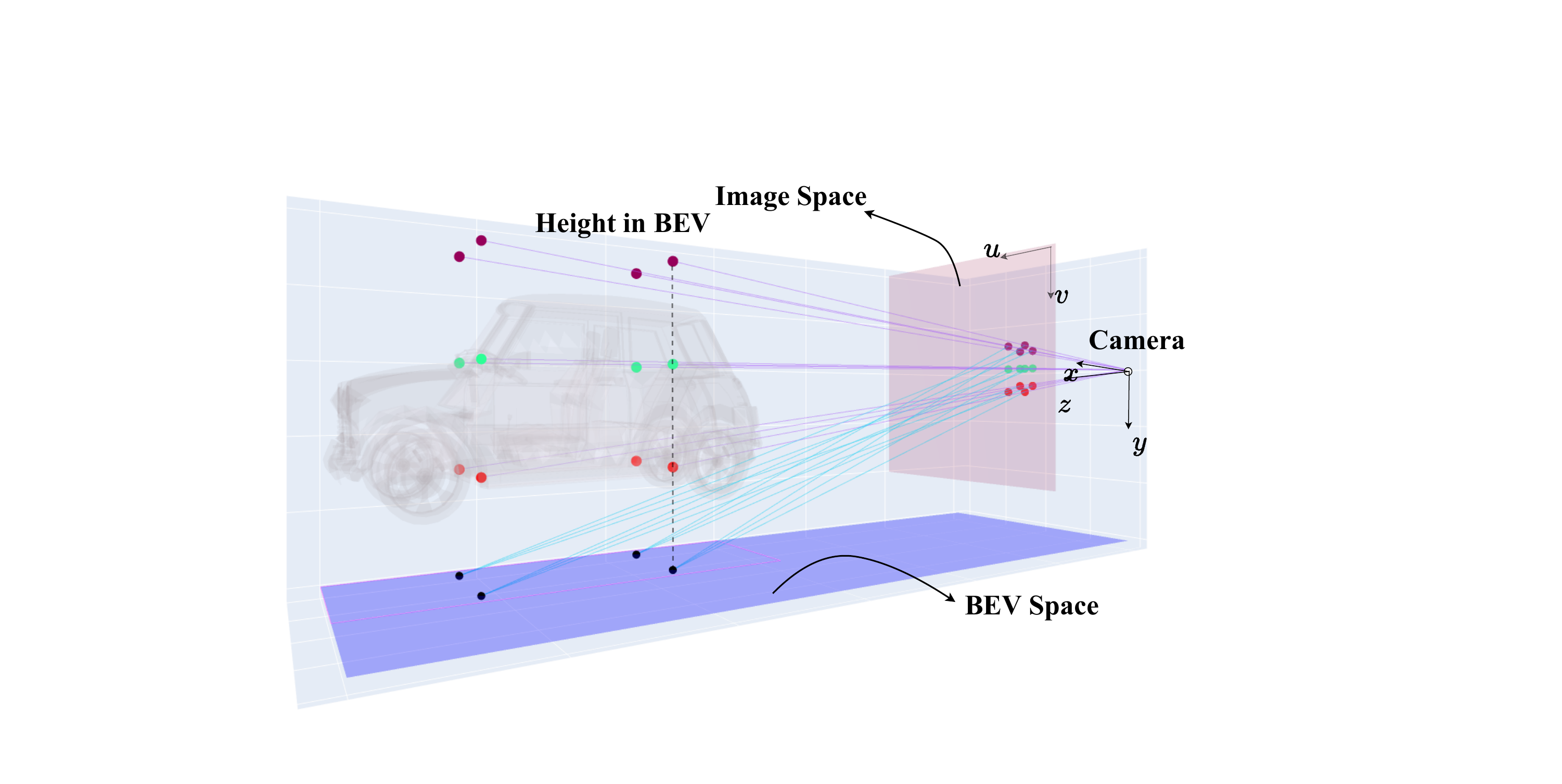}}
    \caption{Two solutions to the ill-posed 2D-3D mapping problem, which are equivalent. In (a), per-pixel depth will be estimated in images, and the feature at a pixel will be projected into a BEV voxel at the depth. In (b), per-grid heights will be estimated or defined in the BEV space. Each anchor is associated with a pixel, and the image feature at that pixel will be accumulated into the grid.}
    \label{fig:main}
\end{figure*}

\section{Method}\label{sec:method}

In this work, we first present a theoretical analysis of depth modeling and height modeling, as they are the basis of 2D to 3D mapping. It could be demonstrated that they are equivalent to achieving mapping from image space to BEV space. Furthermore, we will introduce the pipeline of the proposed HeightFormer. We propose a self-recursive height predictor to estimate heights in BEV and refine them layer by layer. A segmentation-based query mask is also designed to drop background information to improve the efficiency and performance of the model.

\subsection{Theoretical equivalence of depth and height}\label{sec:equivalence}

In this part, we give proof of the equivalence between depth modeling and height modeling in 2D to 3D mapping, as shown in \cref{fig:main}. \uwave{The proof is aimed at showing the feasibility of constructing high-quality BEV features by modeling heights accurately. Besides, at the end of this part, we analyze the advantages of height modeling.}

\paragraph{Precondition} We use two coordinate systems in our work: the image coordinate system and the \uwave{LiDAR coordinate system}. In the image frame, points are noted as $(u,v)$, while in the \uwave{LiDAR frame}$\footnote{Theoretically, there is a camera frame other than the LiDAR frame. Without losing generality, this manuscript does not distinguish between the camera frame and the LiDAR frame, because we assume that the extrinsic parameter matrix 
is an identity matrix.}$ points are noted as $(x,y,z)$. We use $(x,z)$ to denote a point or grid in the \uwave{2D BEV space} for convenience. Given an object in the 3D frame at $(x_{gt}, y_{gt}, z_{gt})$, its coordinate in the image frame is $(u_{gt},v_{gt})$, and its depth is defined as $d_{gt}\triangleq z_{gt}$. These variables satisfy:
\begin{equation}
    \begin{aligned}
        \left[
            \begin{array}{c}u_{gt}\cdot d_{gt} \\ v_{gt} \cdot d_{gt} \\ d_{gt} \end{array}
        \right]      =K
        \left[
            \begin{array}{c}x_{gt} \\ y_{gt} \\ z_{gt} \end{array}
        \right]                \\
        \st
        K  =\left[
            \begin{array}{ccc}
                f_x & 0   & u_0 \\
                0   & f_y & v_0 \\
                0   & 0   & 1
            \end{array}
            \right],
    \end{aligned}
    \label{eq:intrinsic}
\end{equation}
where $f_x,f_y,u_0,v_0$ are intrinsics of the camera. $K$ is the projection matrix that does coordinate transformations.

\paragraph{Assumption} We assume that an object will be correctly detected if its feature is placed close to its ground truth position in the \uwave{2D BEV space}. Mathematically, ``close" is defined as being in the $\epsilon$-neighbourhood of $(x_{gt},z_{gt})$ grid. Here $\epsilon$ is the tolerance of the feature misplacement error. That is to say,
\begin{equation}
    | x - x_{gt} | + | z - z_{gt}| \le \epsilon,\quad \st \epsilon > 0,
    \label{eq:assumption}
\end{equation}
where $(x,z)$ is the coordinate of \uwave{the object's feature in the 2D BEV space}. We simply use the Manhattan distance here to measure the feature misplacement error.

\paragraph{Depth error} Given the depth $d$ of the object, the depth error is $\delta_d=|d-d_{gt}|$. For simplicity, we use absolute error and ignore the relationship between error and depth because the depth of the object is fixed. The 3D position of the object will be modeled as $(x,y,z)$, which follows:
\begin{equation}
    \begin{aligned}
        \left[
            \begin{array}{c}x \\ y \\ z \end{array}
        \right] & =K^{-1}\left[
            \begin{array}{c}u_{gt}\cdot (d_{gt}\pm\delta_d) \\ v_{gt} \cdot (d_{gt}\pm\delta_d) \\ d_{gt}\pm\delta_d \end{array}
        \right]
                .
    \end{aligned}
    \label{eq:depth}
\end{equation}
In this way, the feature of the object will fall at $(x,z)$ in the \uwave{2D BEV space}. Solving \cref{eq:depth} and \cref{eq:assumption},
we can get the upper bound of the depth error$\footnote{Detailed derivation can be found in the appendix.}$:
\begin{equation}
    \delta_{d,max}=\epsilon \cdot \frac{f_x}{|u_{gt}-u_0|+f_x}.
    \label{eq:deptherror}
\end{equation}

\paragraph{Height error} Given a BEV grid at $(x,z)$, the height error is $\delta_y=|y-y_{gt}|$. For all $(x,z)$ in an $\epsilon$-neighbourhood, we can calculate the corresponding sampling location set in the image space. We note this set as $S_{\epsilon}$, which follows:
\begin{equation}
    \begin{aligned}
        S_{\epsilon}\triangleq\Bigg\{ (u,v)^T=&\frac{1}{z}\left[
            \begin{array}{ccc}
                f_x & 0   & u_0 \\
                0   & f_y & v_0 \\
            \end{array}
            \right]\cdot
        \left[ \begin{array}{c}x \\ 
        y
        \\ 
        z \end{array}\right] \\
        \bigg|& \; |x-x_{gt}|+|z-z_{gt}| \le \epsilon \Bigg\}.
    \end{aligned}
\end{equation}
To make sure the mapped feature is correct, $(u_{gt},v_{gt})^T$ must be in $S_\epsilon$. That is to say,
\begin{equation}
    (u_{gt},v_{gt})^T\in S_\epsilon.
    \label{eq:hpe}
\end{equation}
Solving \cref{eq:hpe},
we will get the upper bound of the height error in the $\epsilon$-neighbourhood of $(x_{gt},z_{gt})$, which is$\footnote{Please refer to the appendix.}$:
\begin{equation}
    \delta_{y,max}=
    \epsilon\cdot\frac{|v_{gt}-v_0|}{f_y}\cdot \frac{f_x}{|u_{gt}-u_0|+f_x}.
    \label{eq:heighterror}
\end{equation}

\paragraph{Conclusion}
According to \cref{eq:deptherror}, \cref{eq:heighterror}, and assumption \cref{eq:assumption},
for the given feature misplacement error $\epsilon$,
a depth prediction error of $\delta_{d,max}$ is equivalent to a height prediction error of $\delta_{y,max}$,
because they are both proportional to $\epsilon$. 
\uwave{As a result, considering two models, one models the height, and the other models the depth. Their height or depth estimation error is proportional, and they yield features with similar misplacement errors. Therefore, in terms of error modeling, they are considered equivalent.}

The previous analysis shows the equivalence between two solutions to the ill-posed 2D to 3D mapping problem as well as the feasibility of constructing BEV space by modeling heights. However, although these two solutions are equivalent, height estimation has advantages over depth estimation. \uwave{Height estimation is done in a unified BEV space}, while depth estimation needs to handle the manufacturing differences of cameras. \uwave{We will show the robustness of height-based method with regard to different camera configurations in our experiments. Besides, supervision on height modeling needs no extra data while supervision on depth modeling requires LiDAR information.}

\subsection{Constructing BEV space by height modeling}\label{sec:a2b}

\begin{figure}[bht]
    \subfloat[]{\includegraphics[width=0.48\linewidth]{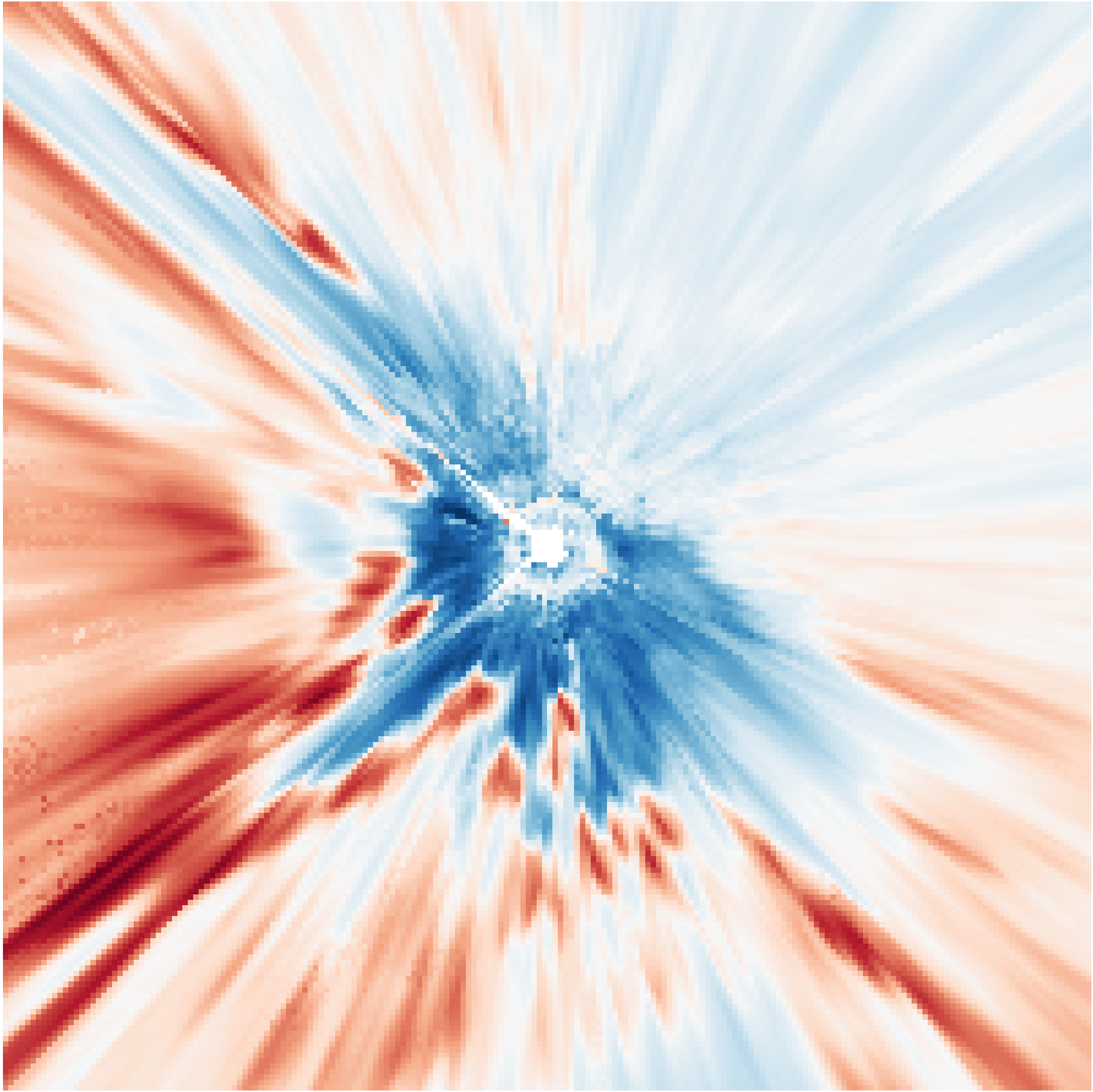}}
    \hfill
    \subfloat[]{\includegraphics[width=0.48\linewidth]{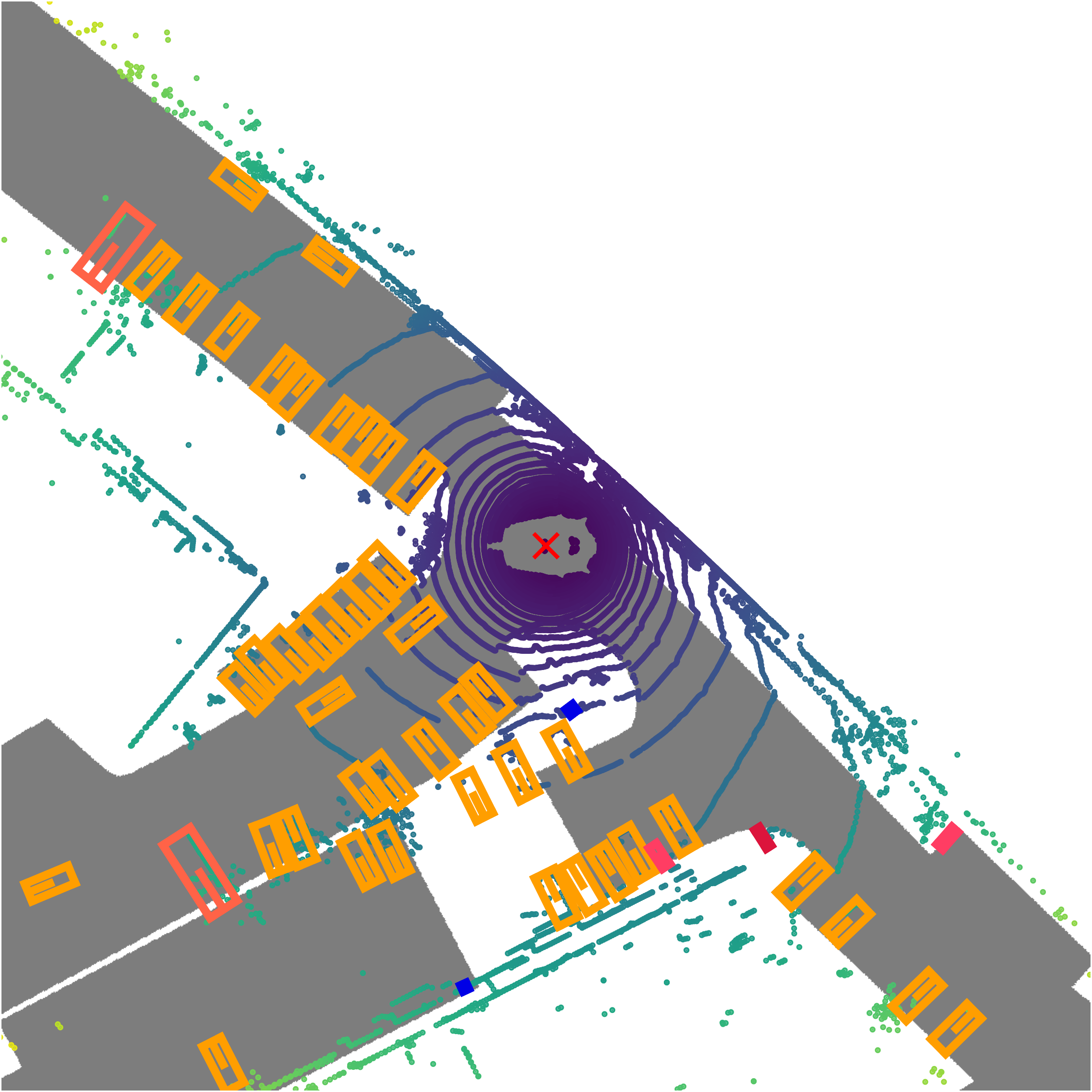}}
    \caption{(a) Heatmap of the weighted heights. (b) The ground truth of bounding boxes under the bird's eye view. Heights are generated by weighting anchor heights with attention weights of spatial cross-attention. The heatmap shows that the information of heights is encoded into attention weights of spatial cross-attention.}
    \label{fig:hlidar}
\end{figure}

Constructing \uwave{BEV features} from image features requires an elaborate feature sampling method. As shown in \cref{fig:main}(b), for a BEV grid, its position and heights can derive several 3D reference points. To acquire the features at these points, they are projected into images, and this step produces 2D reference points. 
BEVFormer\cite{li2022bevformer} aggregates image features into BEV space with spatial cross-attention in its pipeline, and we will follow this practice in our work.

In the vanilla spatial cross-attention, a set of anchor heights are pre-defined in a BEV voxel. The information about heights is implicitly encoded into attention weights. We choose one sample and use the attention weights to weigh those anchor heights. As shown in \cref{fig:hlidar}, the weighted average of anchor heights is structurally similar to the distribution of ground truth bounding boxes.

\begin{figure}[!t]
    \centering
    \includegraphics[width=0.9\linewidth]{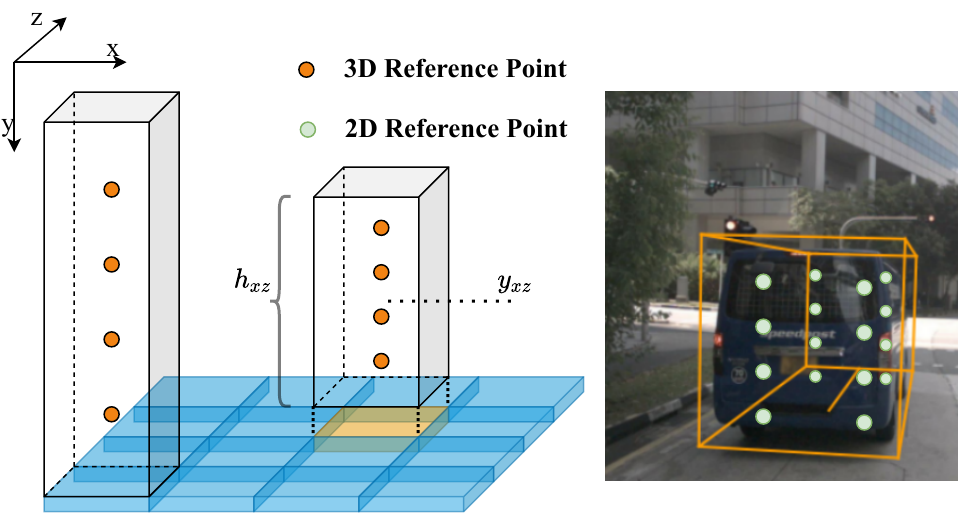}
    \caption{Reference points. 3D reference points are chosen in the range of $[ y_{xz} - h_{xz}/2, y_{xz} + h_{xz}/2 ]$ at equal intervals. They will be projected into multi-view images to generate 2D reference points, and features around those 2D reference points will be gathered into the corresponding BEV queries.}
    \label{fig:anchors}
\end{figure}

Instead of adopting pre-defined anchor heights, we introduce two variables to model the height information about objects and to generate anchor heights. One variable stands for the height of object center $y_{xz}$, and the other stands for the length $h_{xz}$ of the object along the y-axis. As shown in \cref{fig:anchors} the anchor heights will be uniformly distributed in the range of $[ y_{xz} - h_{xz}/2, y_{xz} + h_{xz}/2 ]$.

As the variance of heights could be large, we introduce uncertainties to model the distribution of heights and optimize heights and uncertainties jointly. For convenience, we use an indicator $I_{x,z}$ to indicate if the grid at $(x,z)$ has any object in it. The heights and uncertainties are jointly supervised as follows:
\begin{equation}
    \begin{aligned}
        L_h(x,z) & =\left\{\begin{array}{cc}
                               \sqrt{2}\ \cfrac{|h_{xz}-h_{xz}^{gt}|}{\sigma_{h,xz}}+\log\sigma_{h,xz}, & I_{xz}=1 \\
                               \cfrac{1}{\sigma_{h,xz}},                               & I_{xz}=0
                           \end{array}\right.
    \end{aligned}
    \label{eq:loss0}
\end{equation}
where $\sigma_h$ indicates the uncertainty of the estimated height. This loss function adopts the Laplace prior\cite{chen2020monopair,kendall2017uncertainties} on the distribution of heights, which is modeled as $p(h)=\frac{1}{2\sigma_h}\exp(-\frac{|h-h_{gt}|}{\sigma_h})$. The loss function $L_y(x,z)$ for $y$ is similar to \cref{eq:loss0}.

\subsection{Ground truth of heights in BEV}

\begin{figure}[bht]
    \centering
\includegraphics[width=0.9\linewidth, trim=0 0 0 0, clip]{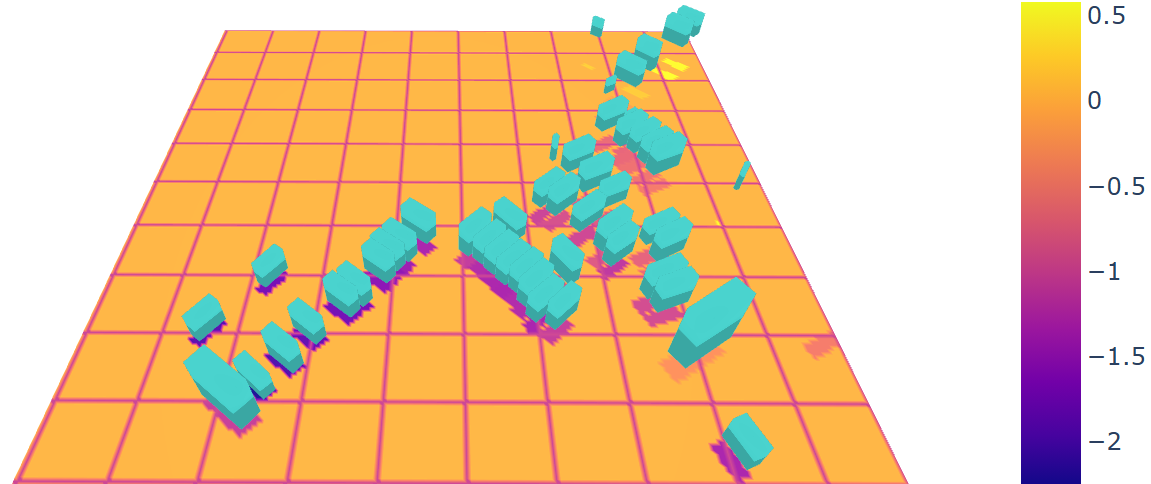}
    \caption{Demo of the ground truth heights $y_{xz}$. The blue cubes are boundings boxes of objects. For the effect of visualization, they are stretched along the height dimension.}
    \label{fig:gt}
\end{figure}

Different from depth supervision which requires LiDAR information, height supervision is easy to acquire and no extra data is directly required. In this work, we project the bounding boxes of objects into the 2D BEV space to format ground truth heights. As the height of the object center, $y_{xz}$ is acquired from the center coordinates of the bounding box in the BEV frame. The object height $h_{xz}$ is acquired from the size of the object. If multiple grids represent the same object, they will share the same heights. In this way, each object is represented with a planar quadrilateral in height maps. An example of the ground truth heights is shown in \cref{fig:gt}. 
Bounding boxes marked as blue are projected into the BEV space, and the height of object center $y_{xz}$ is plotted in a heatmap. The range of $y_{xz}$ is from -5m to 3m in the LiDAR coordinate system.

\subsection{Network design}
\begin{figure}[bht]
    \centering
    \includegraphics[width=\linewidth]{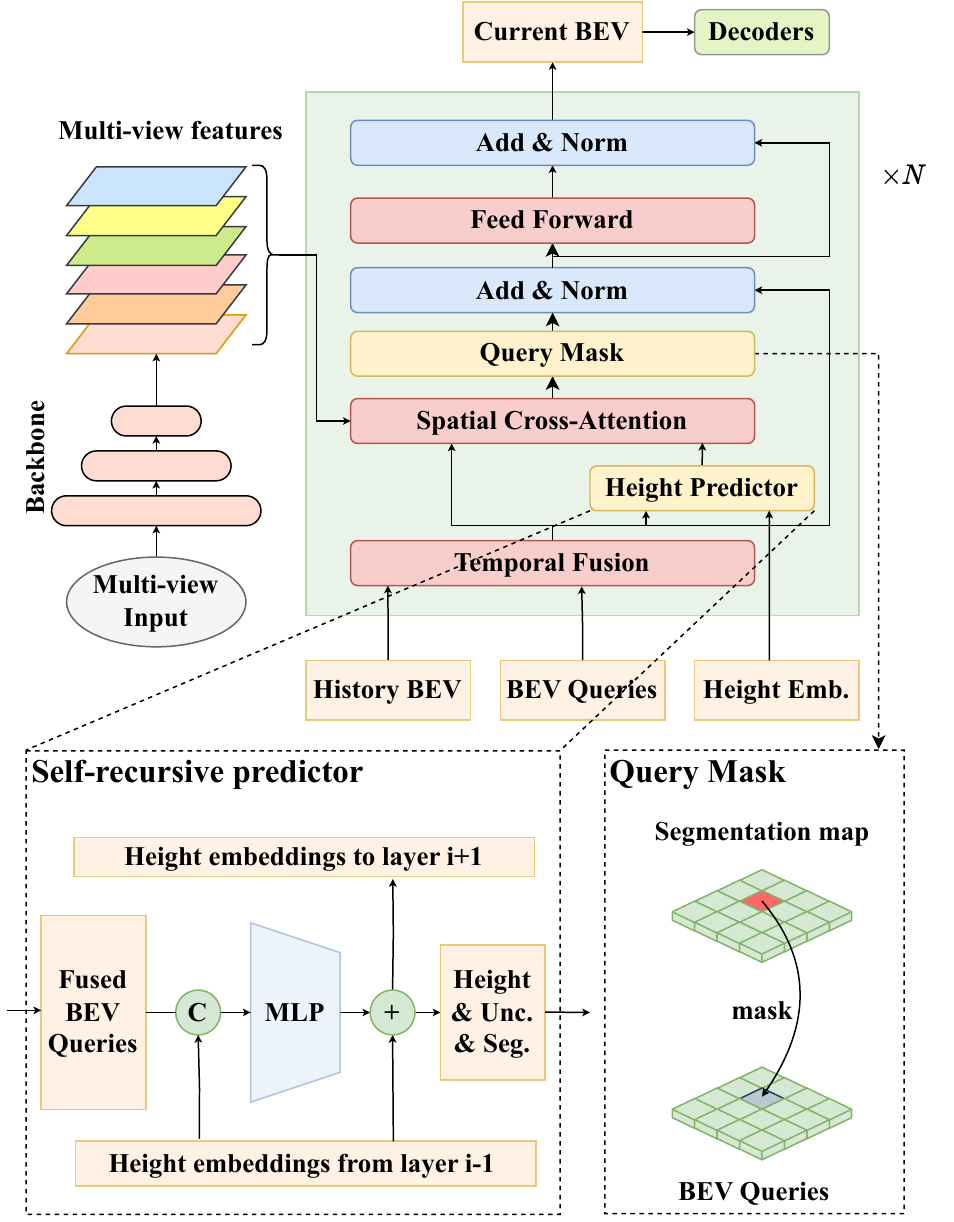}
    \caption{The main architecture of HeightFormer.}
    \label{fig:arch}
\end{figure}

The architecture of the proposed HeightFormer is shown in \cref{fig:arch}. We mainly introduce two components: the self-recursive height predictor and the segmentation-based query mask. Other modules will be introduced briefly.

\paragraph{Self-recursive height predictor} As the BEV encoder is composed of several successive layers, we design a self-recursive height predictor to refine heights layer by layer and use height embeddings to preserve height information. The predictor takes height embeddings of the previous layer and BEV queries from the temporal fusion module as input and outputs height embeddings, which will be normalized to get the final heights and uncertainties. This process is noted as:
\begin{equation}
    E_l=f_l(f_{l-1}(\cdots f_1(E_0,Q_1)\cdots,Q_{l-1}),Q_{l}),
\end{equation}
where $E_l$ means height embeddings of layer $l$ and $E_0$ means initial height embeddings. $f_l(\cdot)$ stands for the predictor of layer $l$. $Q_l$ stands for BEV queries from the temporal fusion module of layer $l$.

\paragraph{Initial height embeddings} \uwave{are designed to make initial anchor heights uniformly distributed from the ground to the sky. For a BEV gird at $(x,z)$, its corresponding height embedding is initialized as $(0.5, 1.0, 0, 0, 0)$. The elements stand for $y$, $h$, $\log\sigma_y$, $\log\sigma_h$, $logit$. The last element $logit$ is used for segmentation.}

\paragraph{Segmentation-based query mask} \uwave{After statistical analysis, it is found that many queries do not contain any annotated object. For these queries, simply defining low heights may lead to inconsistent results for the ground, buildings, trees and other unannotated objects. As a result, the predicted heights of these queries are not constrained.} To avoid gathering irrelevant features from the background, we introduce the segmentation-based query mask. The self-recursive height predictor will predict an extra binary segmentation map, which indicates the probability of covering an object. The gathered features in queries that have low probability will be filtered.

\paragraph{Other modules} Temporal fusion modules\cite{li2022bevformer} \uwave{fuse information from previous frames with Self Attention. In this module, the query is the current BEV query, and the value is the concatenated (previous and current) BEV query.} DETR-style decoders are adopted to detect objects based on BEV features.
\section{Experiments}

\subsection{Experimental details}
\paragraph{Dataset} The proposed method is evaluated on the NuScenes\cite{nuscenes2019} detection dataset.
As a challenging dataset for autonomous driving, NuScenes provides data of 1k scenes, from a sensor suite that consists of 6 cameras, 1 LiDAR, 5 RADAR, 1 GPS, and 1 IMU. The {\it train} set contains 28,130 samples and the \uwave{{\it validation} set} contains 6,019 samples.

\paragraph{Metrics} The main metrics that reflect the performance are mAP and NDS. mAP is widely used in detection tasks. NDS stands for NuScenes Detection Score, which consists of six sub-metrics: mAP, mATE, mASE, mAOE, mAVE, and mAAE. NDS is computed as:
\begin{equation}
    \text{NDS}=\cfrac{1}{10}[5\text{mAP}+\sum_{\text{mTP}\in \mathbb{TP}}[1-\min(1, \text{mTP})],
    \label{eq:nds}
\end{equation}
where $\mathbb{TP}$ is the set of sub-metrics expect mAP. mATE, mASE, and mAOE measure the localization error, the scale error, and the orientation error respectively. mAVE measures velocity prediction error. mAAE measures the error of object classification. For NDS and mAP, higher is better. For other sub-metrics, the opposite is true.

\paragraph{Settings} As our model is built based on BEVFormer\cite{li2022bevformer}, we share most settings with BEVFormer. \uwave{For time fusion module, 4 history BEV queries are used in the {\it base} setting and 3 BEV queries are used in the {\it tiny} setting.} The BEV space is divided into 200$\times$200 grids in both {\it base} and {\it tiny} settings, with each grid standing for an area of 0.512 meters by 0.512 meters. We supervise the height predictor of the last encoder layer with the loss function stated in \cref{eq:loss0}. The extra segmentation map is supervised with binary focal loss\cite{lin2017focal}. A position-aware loss weight called BEV centerness\cite{xie2022m} which pays more attention to distant areas is also used to improve the height estimation accuracy of distant grids.

\subsection{Ablation study}

\paragraph{Performance upper bound} We first replace the height predictors with ground truth heights to evaluate whether accurate height estimation could improve the performance of detection, and we call this method HeightFormer-gt. For BEV girds containing objects, predicted heights are replaced with ground truth heights. For others, predicted heights are replaced with $y_{xz}=0.5$ and $h_{xz}=1.0$ (normalized), which degrades into the situation of BEVFormer. Results are shown in \cref{tab:expa0}.

\begin{table}[hbt]
    \centering
    \caption{The upper bound of performance. The anchor heights of HeightFormer-gt are uniformly distributed in the range of $[y_{gt} - h_{gt}/2, y_{gt} + h_{gt}/2 ]$. *: The experiments about Lift-Splat\cite{philion2020lift} is done by BEVDepth\cite{li2022bevdepth}.}
    \begin{tabular}{lccc}
        \toprule
        Model             & Condition & NDS$\uparrow$ & mAP$\uparrow$ \\
        \midrule
        Lift-Splat$^*$    & -         & 0.327         & 0.282         \\
        Lift-Splat-gt$^*$ & depth     & {\bf 0.515}         & {\bf 0.470}         \\
        \midrule
        BEVFormer         & -         & 0.517         & 0.416         \\
        HeightFormer-gt      & height    & {\bf 0.725}   & {\bf 0.789}   \\
        \bottomrule
    \end{tabular}
    \label{tab:expa0}
\end{table}

As shown in \cref{tab:expa0}, HeightFormer-gt exceeds BEVFormer by 0.208 in NDS and 0.374 in mAP, which can be considered as the performance upper bound. 
\uwave{However, in reality, height or depth estimation error could be amplified when considering the final detection error. As a result, it is not feasible to achieve this theoretical upper bound.}
In the meanwhile, Lift-Splat\cite{philion2020lift} in \cref{tab:expa0} is a method that constructs the BEV space with depth modeling and according to BEVDepth\cite{li2022bevdepth} introducing ground truth depth improves mAP of the Lift-Splat style detector by 0.188. The experiments in \cref{tab:expa0} show that there is room for improvement whether in depth modeling or height modeling.

Besides, it should be noted that the two types of methods, LSS and BEVFormer, cannot be compared directly as they have different network architectures. \uwave{We list these results here to show the potential of height-based methods.}

\paragraph{Effectiveness of explicit height modeling} We present the vanilla HeightFormer which turns fixed reference points into adaptive reference points. The adaptive reference points are generated with predicted anchor heights. In this way, height information is explicitly learned by the height predictor. To show the effectiveness of explicit height modeling, it is compared with BEVFormer which encodes height information into attention weights of spatial cross-attention in an implicit way. The vanilla HeightFormer simply has a multilayer perceptron as the height predictor. Results are shown in \cref{tab:expa1}. \uwave{Explicit height modeling brings a gain of 0.5 percentage points in both NDS and mAP for the ``base'' model. For the ``tiny'' model, the improvements come to 1.8 and 1.1 percentage points.
}


\begin{table}[hbt]
    \centering
    \caption{Ablation study on height modeling. *: Vanilla HeightFormer which predicts heights in a standalone way. $\dag$: Official results, no history frames.}
    \begin{tabular}{lcccc}
        \toprule
        Model    & Config  & Height   & NDS$\uparrow$ & mAP$\uparrow$ \\
        \midrule
        BEVFormer & Base & implicit & 0.517         & 0.416         \\
        HeightFormer* & Base & explicit & {\bf 0.522}         & {\bf 0.421}     \\
        \midrule
        BEVFormer     & Tiny & implicit & 0.403         & 0.288         \\
        HeightFormer* & Tiny & explicit & {\bf 0.421}   & {\bf 0.299}     \\
        \bottomrule
    \end{tabular}
    \label{tab:expa1}
\end{table}

\paragraph{Effectiveness of the network design}
Based on the vanilla HeightFormer which predicts heights standalone in each layer, height embeddings are added to formulate self-recursive predictors. For BEV grids that do not cover any object, the query mask is applied. Two types of masks are designed for comparison. The uncertainty-based mask filters BEV queries that have high uncertainties of the predicted heights. The segmentation-based mask filters BEV queries that are less likely to have objects. Results are shown in \cref{tab:expa3}.

The self-recursive way of predicting heights brings an improvement of 0.3 percentage points in NDS. Besides, we observe that there is a giant gap between the outputs of successive predictors when they are standalone, and refining the heights layer by layer in a self-recursive way mitigates this issue. 

The two types of masks both improve the NDS, while the segmentation-based mask brings a gain of 0.5  percentage points in mAP. The improvement mainly comes from filtering irrelevant features and introducing segmentation as an auxiliary task.
In the proposed method, there is no proper way to define the heights of BEV grids that do not cover any object. As a result, the predicted heights in these grids are not explainable and the sampled features are from irrelevant background areas. Applying a mask mitigates this issue.

\begin{table}[hbt]
    \centering
    \caption{Ablation study on the effectiveness of the network design. SR: a self-recursive way of predicting heights. Unc.M: uncertainty-based mask. Seg.M: segmentation-based mask.}
    \begin{tabular}{ccc|cc}
        \toprule
        SR         & Unc.M      & Seg.M      & NDS$\uparrow$ & mAP$\uparrow$ \\
        \midrule
                   &            &            & 0.522         & 0.421         \\
        \checkmark &            &            & 0.525         & 0.422         \\
        \checkmark & \checkmark &            & {\bf 0.528}   & 0.424         \\
        \checkmark &            & \checkmark & 0.527         & {\bf 0.427}   \\
        \bottomrule
    \end{tabular}
    \label{tab:expa3}
\end{table}

\paragraph{Robustness of height modeling} 
As described in Introduction, although both depth and height modeling can provide an extra condition for 2D to 3D mapping, height modeling has a unique advantage in that it can process any camera rig. This is because estimating depth in the image space might be influenced by the camera rigs, but estimating height is simple and robust since all information has been mapped to the unified BEV space. \uwave{To verify this property, we test our method with depth modeling specifically with the back camera, which has a different focal length from other cameras.} We use BEVDepth\cite{li2022bevdepth} as the depth modeling method. Results are shown in \cref{tab:expa4}.

\begin{table}[htb]
    \centering
    \caption{Ablation study on the robustness of height modeling. Although the depth modeling method has higher overall performance (due to extra LIDAR data), our method still achieves higher performance on the back camera which has a different focal length and camera configuration. This shows the robustness of height modeling.}
    \begin{tabular}{c|cc|cc}
        \toprule
        \multirow{2}{*}{Extra condition}
                  & \multicolumn{2}{c|}{Overall} & \multicolumn{2}{c}{Back Cam}  \\
         & mAP & NDS & mAP & NDS  \\
        \midrule
        Depth modeling   & {\bf 0.330}  &  {\bf  0.436}   & 0.273  &   0.398             \\
        Height modeling & 0.299     & 0.421 & {\bf 0.279}   &  {\bf  0.413}  \\
        \bottomrule
    \end{tabular}
    \label{tab:expa4}
\end{table}

From \cref{tab:expa4} we can see that although BEVDepth has higher overall performance \uwave{(due to extra LiDAR data training and different model architectures)}, height modeling's performance on the back camera is higher, which highlighted a great performance gap between the back camera and other cameras for BEVDepth. This means height modeling is more robust to different camera rigs, even when the cameras' focal lengths are different.

\subsection{LiDAR supervision}
\begin{figure}[bht]
    \centering
    \includegraphics[width=0.9\linewidth, trim=20 30 20 40, clip]{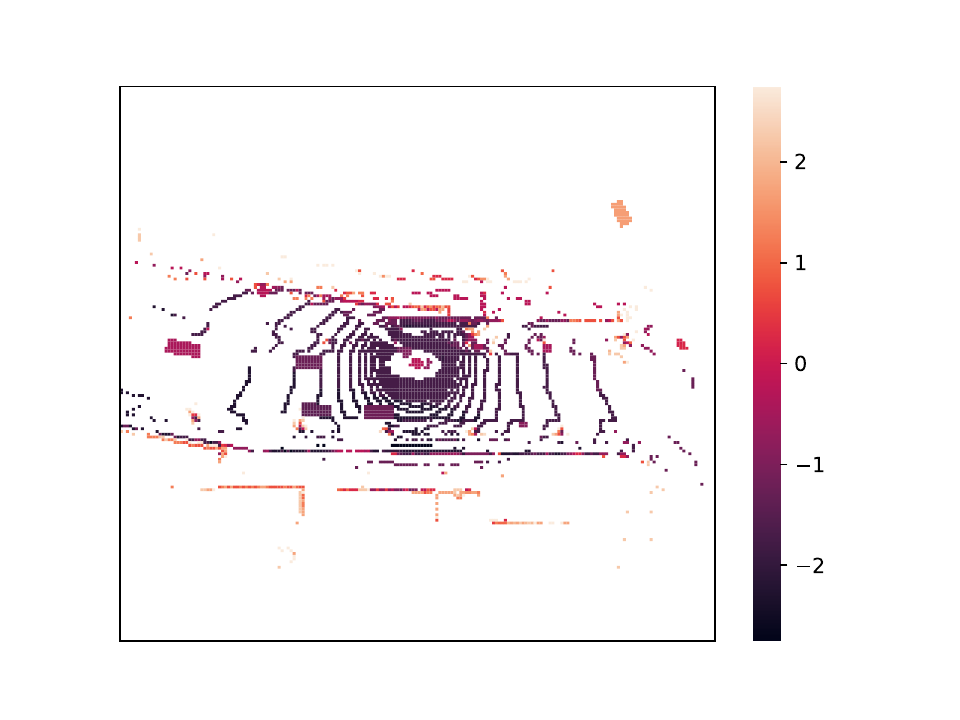}
    \caption{Demo of fused heights. This figure only shows $y$. LiDAR points are projected into BEV grids with their heights: (a) The height interval that contains the most points is used to formulate $y_{xz}$; (b) The lowest point in a grid is used to formulate $y_{xz} - h_{xz}/2$.}
    \label{fig:fused}
\end{figure}

 Ground truth heights are critical for the HeightFormer module. However, with the bounding boxes of objects, we can only obtain the heights of grids that have objects, which limits the performance of the proposed method. To show how much further improvement can be achieved with extra LiDAR supervision, we introduce LiDAR supervision in this part.

First, LiDAR points are projected into BEV grids with the height range meshed into 16 intervals. For each BEV grid, its height $y_{xz}$ is the height of the interval containing the most points. Its lower bound is the height of the lowest point, which formulates $y_{xz} - h_{xz} / 2$. Furthermore, heights from LiDAR points and heights from bounding boxes are fused to formulate the ground truth heights. For each BEV gird, the height from ground truth bounding boxes is preferred. The fused heights serving as the ground truth heights are shown in \cref{fig:fused}.

With the extra supervision of LiDAR information, the performance is improved by 0.7 percentage points. The ablation study is shown in \cref{tab:explidar}. 

\begin{table}[htb]
    \centering
    \caption{Ablation study on LiDAR supervision. The models here take ResNet50 as the backbone and the BEV space is meshed into 200 grids by 200 grids. }
    \begin{tabular}{lccc}
        \toprule
        Model     & Height           & NDS$\uparrow$ & mAP$\uparrow$ \\
        \midrule
        BEVFormer & implicit         & 0.403         & 0.288         \\
        HeightFormer & explicit         & 0.421         & 0.299         \\
        HeightFormer & explicit + lidar & {\bf 0.428}         & {\bf 0.307}         \\
        \bottomrule
    \end{tabular}
    \label{tab:explidar}
\end{table}

We can note the extra LiDAR supervision brings little improvement to the performance of the proposed method compared with the improvement brought by explicit height modeling. \uwave{A potential reason is that LiDAR points are sparse and occluded positions are not covered. This might yield inconsistent supervision information. This is different from the situation of depth modeling, in which most pixels can be paired with one or more LiDAR points. However, the advantage of our method is that it is a cost-free approach to improve performance and requires no extra data.}

\subsection{Generalization ability}
\begin{figure}[hbt]
    \centering
    \includegraphics[width=0.9\linewidth]{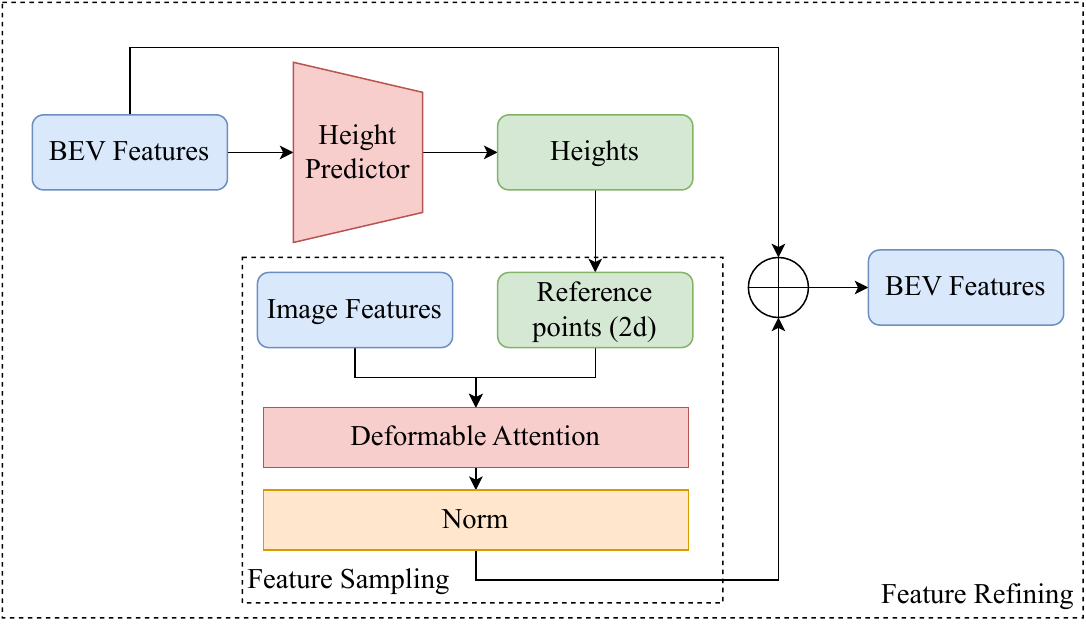}
    \caption{The pipeline of the HeightFormer plugin. HeightFormer can serve as a plugin to refine other types of BEV representations. This plugin takes image features and BEV features as the input and outputs the refined BEV features.}
    \label{fig:refine}
\end{figure}

In this part, we will show that the proposed HeightFormer can also serve as a plugin to refine other types of BEV representations. The pipeline is shown in \cref{fig:refine}.

Taking BEVDepth as an example, it lifts image features with predicted depth distribution into voxel features and splats voxel features into BEV features with voxel pooling. Here we make HeightFormer a plugin of BEVDepth and improve the performance of BEVDepth with the plugin. The results are shown in \cref{tab:expgen}.

\begin{table}[htb]
    \centering
    \caption{Performance of height-based BEV feature refinement. The baseline models both take ResNet50 as the backbone.}
    \begin{tabular}{lccc}
        \toprule
        Model     & Refinement & NDS$\uparrow$ & mAP$\uparrow$ \\
        \midrule
        BEVFormer &            & 0.403         & 0.288         \\
        BEVFormer & 3 layers & {\bf 0.421}         & {\bf 0.299}         \\
        \midrule
        BEVDepth  &            & 0.366         & {\bf 0.273}             \\
        BEVDepth  & 1 layer & {\bf 0.369}         & {\bf 0.273}             \\
        \bottomrule
    \end{tabular}
    \label{tab:expgen}
\end{table}

We first train a BEVDepth model without BEV data augmentation for 20 epochs. Its best performance is 0.366 in NDS. We insert a single layer of the HeightFormer plugin into BEVDepth and train the new model with the same settings. The new model's best performance is 0.369 in NDS. The NDS is improved by 0.3 percentage points, which shows that the HeightFormer plugin effectively improves the BEV representations and thus improves the performance.

\subsection{Qualitative Results}

\begin{figure}[htb]
    \centering
    \includegraphics[width=0.85\linewidth]{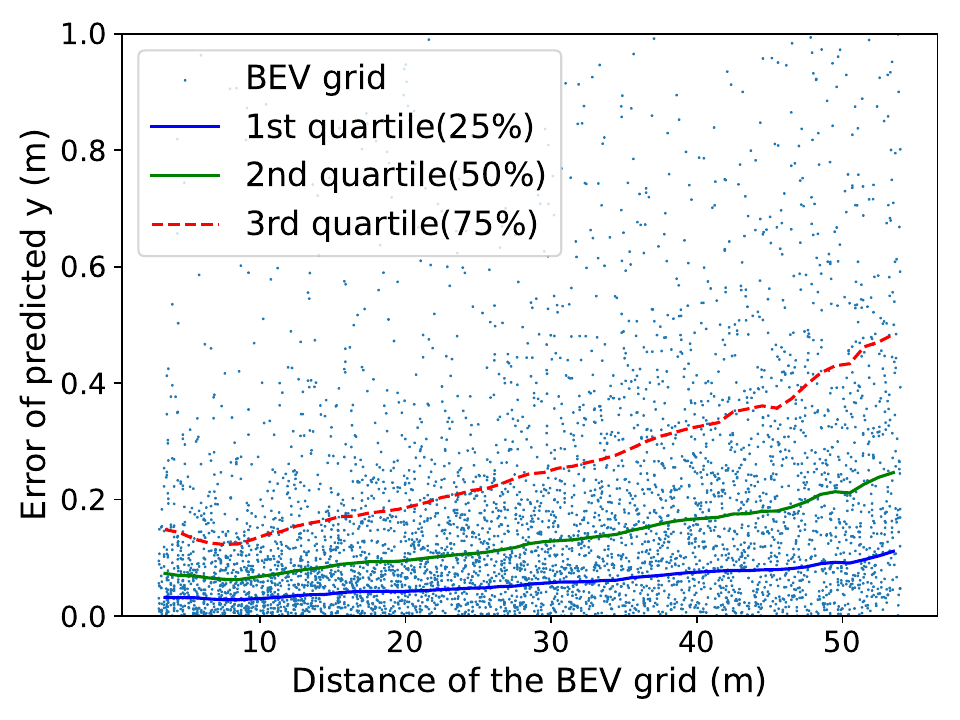}
    \caption{The relationship between prediction error of $y$ and distances. Heights are predicted by HeightFormer-{\it tiny}. Distant BEV grids tend to have higher height prediction errors.}
    \label{fig:hpe}
\end{figure}

\paragraph{Height prediction error} The performance of height predictors is key to the process of gathering features. To show that heights could be estimated well, we plot the prediction error of $y$ in \cref{fig:hpe}. The data points in the figure stand for random-chosen BEV grids of all samples. For close BEV grids, 75\% of them have an error of less than 0.2m. For distant BEV grids, 75\% of them have an error of less than 0.5m. This level of error will keep most reference points falling on objects.

\begin{figure}[hbt]
    \centering
    \includegraphics[width=1.0\linewidth]{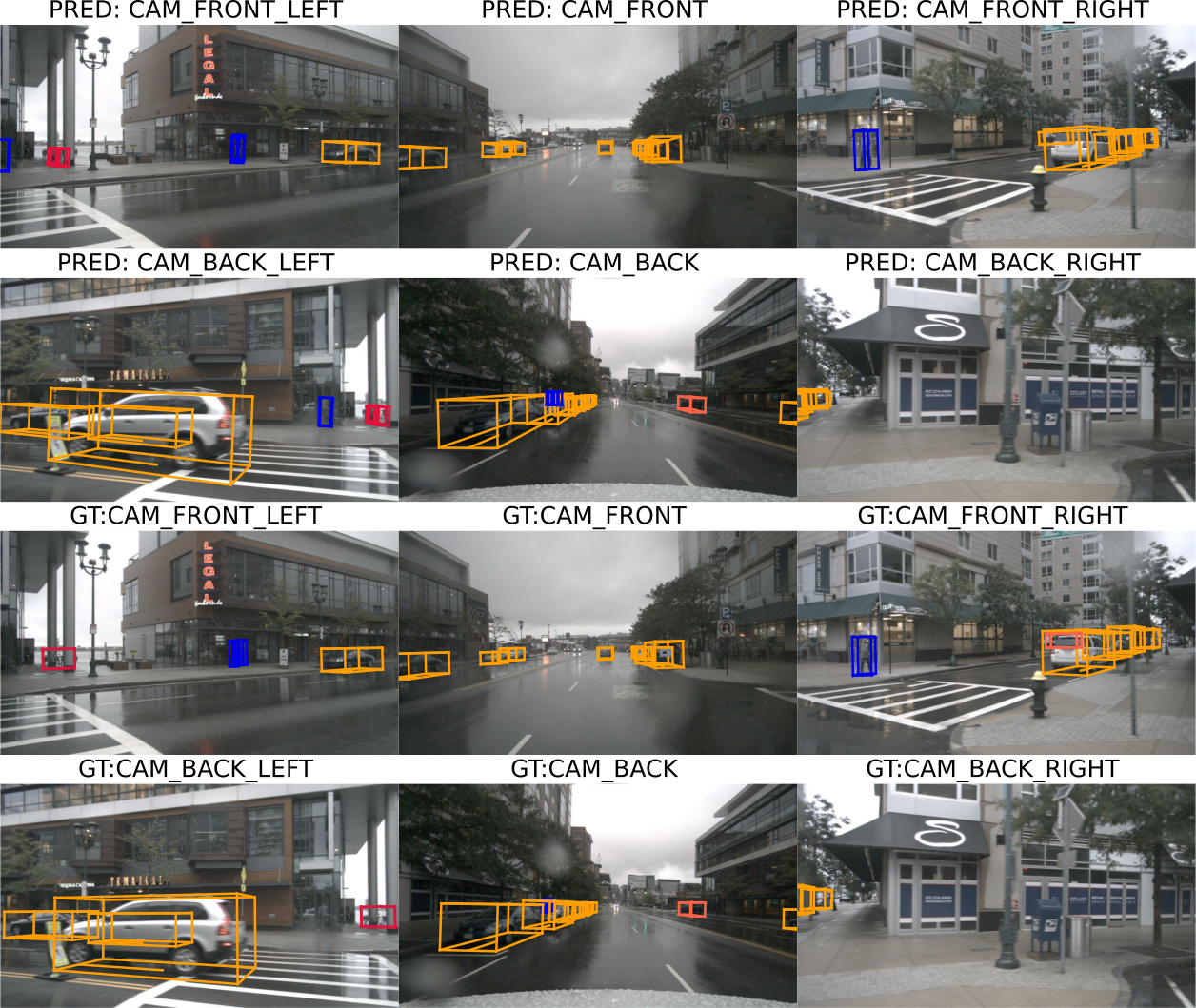}
    \caption{Visualization of predicted bounding boxes. The first two rows are predicted results, and the last two rows are the ground truth. Different bounding box colors indicate different object categories.}
    \label{fig:det}
\end{figure}

\paragraph{Visualization of bounding boxes} In this part, we take a glance at the detected bounding boxes. \cref{fig:det} shows the detection results of one sample by HeightFormer-{\it tiny}. In this sample, most objects have been correctly detected, while a few distant or occluded objects are misclassified.  
This is because distant objects occupy few pixels in images and occluded objects' features cannot be sampled.

\subsection{Benchmark results}

\paragraph{NuScenes \uwave{{\it validation} set}}
We report the results on NuScenes \uwave{{\it validation} set} in \cref{tab:exp1} and compare the proposed HeightFormer with state-of-the-art camera-only methods.
Compared with BEVFormer, HeightFormer improves the NDS by 1.5 percentage points.
Compared with the SOTA BEVDepth which introduces LiDAR in training and takes data augmentation in both image space and BEV space, the proposed HeightFormer has one percentage point of improvement in mAP, which is the key metric of detection.

{{\color{blue}}
After a detailed analysis of all sub-metrics, we find that the improvement of mAP mainly comes from AP at small thresholds, for example, AP@1.0m. In the meanwhile, the improvement in the detection of rare classes is larger than that of common classes: the improvements of AP about ``bus", ``trailer" and ``motorcycle" are larger than that of ``car".
A potential reason is that the proposed query mask filters out the background and thus reduces false positives. This is helpful for the detection of rare classes.}

\begin{table}[bth]
    \centering
    \caption{3D detection results on NuScenes \uwave{{\it validation} set}. The listed models mostly take R101-DCN as the backbone and require no extra LiDAR data except for BEVDepth. $\dag$: Trained with CBGS\cite{zhu2019class}. *: Take LiDAR as auxiliary information in the training phase.
    }
    \begin{tabular}{l|c|cc}
        \toprule
        Model                             & Backbone & NDS$\uparrow$ & mAP$\uparrow$ \\
        \midrule
        BEVDepth$^*$\cite{li2022bevdepth} & R101-DCN & 0.538         & 0.419         \\
        \midrule
        FCOS3D\cite{wang2021fcos3d}       & R101-DCN & 0.372         & 0.295         \\
        DETR3D\dag\cite{wang2022detr3d}   & R101-DCN & 0.434         & 0.349         \\
        PGD\cite{wang2022probabilistic}   & R101-DCN & 0.428         & 0.369         \\
        BEVDet\dag\cite{huang2021bevdet}  & Swin-T   & 0.472         & 0.393         \\
        PolarDETR-T\cite{chen2022polar}   & R101-DCN & 0.488         & 0.383         \\
        UVTR\cite{li2022unifying}         & R101-DCN & 0.483         & 0.379         \\
        PETR\dag\cite{liu2022petr}        & R101-DCN & 0.442         & 0.370         \\
        Ego3RT\cite{lu2022learning}       & R101-DCN & 0.450         & 0.375         \\
        BEVFormer\cite{li2022bevformer}   & R101-DCN & 0.517         & 0.416         \\
        Ours                         & R101-DCN & {\bf 0.532}   & {\bf 0.429}   \\
        \bottomrule
    \end{tabular}
    \label{tab:exp1}
\end{table}

\begin{table}[htb]
    \centering
    \caption{3D detection results on NuScenes {\it test} set. The listed models are all camera-only methods. The proposed HeightFormer is trained without tricks. *: Extra LiDAR supervision.}
    \begin{tabular}{l|c|c|c}
        \toprule
        Model                           & Backbone & NDS$\uparrow$ & mAP$\uparrow$ \\
        \midrule
        BEVDepth$^*$\cite{li2022bevdepth}   & V2-99    & 0.600         & 0.503 \\
        \midrule
        DD3D\cite{park2021pseudo}       & V2-99    & 0.477         & 0.418         \\
        BEVDet\cite{huang2021bevdet}    & V2-99    & 0.488         & 0.424         \\
        DETR3D\cite{wang2022detr3d}     & V2-99    & 0.479         & 0.412         \\
        UVTR\cite{li2022unifying}       & V2-99    & 0.551         & 0.472         \\
        PETR\cite{liu2022petr}          & V2-99    & 0.504         & 0.441         \\
        Ego3RT\cite{lu2022learning}     & V2-99    & 0.473         & 0.425         \\
        BEVFormer\cite{li2022bevformer} & V2-99    & 0.569         & 0.481         \\
        HeightFormer                       & V2-99    & {\bf 0.573}   & {\bf 0.481}   \\
        \bottomrule
    \end{tabular}
    \label{tab:exptest}
\end{table}

\paragraph{NuScenes {\it test} set}
We report the results on NuScenes {\it test} set in \cref{tab:exptest}. The listed models all take VoVNet (V2-99)\cite{lee2019energy} initialized from DD3D\cite{park2021pseudo} as the backbone. The proposed HeightFormer improves the NDS of BEVFormer by 0.4\%. \uwave{However, this improvement is not significant compared with the improvement on the {\it validation} set. A potential reason is that the commonly adopted VoVNet backbone is pre-trained on a depth estimation task, which does not bring much benefit to height modeling.}

\section{Conclusion and Limitation}
In this work, we analyze the 2D to 3D mapping problem in constructing the BEV space and give proof of the equivalence between depth estimation in the image space and height estimation in the BEV space. Based on the proof, {{\color{blue}}
we propose HeightFormer which explicitly models heights in BEV without extra LiDAR supervision for car-side situations.} According to our experiments, the proposed self-recursive height predictor can model heights accurately, and the segmentation-based query mask effectively improves the performance of detection. We also show that height modeling is more robust to different camera rigs compared to depth modeling.

However, there is a limitation in this work. The ground truth heights are acquired by projecting bounding boxes, meaning only a few queries have heights defined. Even if we introduce LiDAR supervision, most grids have no LiDAR points falling in them. As a result, the heights of queries at these grids are still learned implicitly. To mitigate this issue, we filter these queries to avoid introducing irrelevant features in the sampling procedure. \uwave{A new way of utilizing LiDAR for height supervision needs to be researched in the future.}

{\appendix

\section*{Detailed derivation of $\delta_{d,max}$}

According to the analysis of depth prediction error in \cref{sec:equivalence}, if an object is correctly detected, its projected position $(x,z)$ in the BEV space should fall in the $\epsilon$-neighbourhood of its ground truth position $(x_{gt},z_{gt})$. That is to say:
    \begin{align}
        |x-x_{gt}|&+|z-z_{gt}| \le \epsilon, \label{eq:d1}\\
        \st 
            \left[
                \begin{array}{c}x \\ y \\ z \end{array}
            \right] &=K^{-1}\left[
                \begin{array}{c}u_{gt}\cdot d \\ v_{gt} \cdot d \\ d \end{array}
            \right], \label{eq:d2}\\
            \left[
                \begin{array}{c}x_{gt} \\ y_{gt} \\ z_{gt} \end{array}
            \right]   &=K^{-1}\left[
                \begin{array}{c}u_{gt}\cdot d_{gt} \\ v_{gt} \cdot d_{gt} \\ d_{gt} \end{array}
            \right] \label{eq:d3} \\
        \label{eq:d4}
    \end{align}
where $(u_{gt}, v_{gt})$ is the position of the object in the image.
$d$ is the given depth at $(u_{gt}, v_{gt})$. The feature at $(u_{gt}, v_{gt})$ is gathered into the BEV grid at $(x,z)$. $K$ is the camera's intrinsic matrix which does the transformation between the image frame and the BEV frame as stated in \cref{eq:intrinsic}.$\footnote{Theoretically, there is a camera frame other than the BEV frame. Without losing generality, this manuscript does not distinguish between the camera frame and the BEV frame, because we assume that the extrinsic parameter matrix 
is an identity matrix.}$
Expanding \cref{eq:d2}, \cref{eq:d3} and \cref{eq:d4}, and we get
\begin{align}
    x-x_{gt} &= \frac{1}{f_x}(u_{gt}-u_0)(d-d_{gt}),                      \\
    z-z_{gt} &= d-d_{gt}.
\end{align}
Substituting them into \cref{eq:d1}, and we can solve out the upper bound of $|d-d_{gt}|$ as follows:
\begin{equation}
    \delta_{d,max}=\epsilon \cdot \frac{f_x}{|u_{gt}-u_0|+f_x},
\end{equation}
which is described in the \cref{eq:deptherror}.

\section*{Detailed derivation of $\delta_{y,max}$}

\label{sec:intro}
According to the analysis of height prediction error in \cref{sec:equivalence}, if an object is correctly detected, the sampling locations should cover the ground truth position of the object's feature. This is concluded as:
\begin{equation}
    \begin{aligned}
        (u_{gt},&v_{gt})^T\in S_{\epsilon},                                  \\
        \st\; S_{\epsilon}\triangleq\Bigg\{ (u,v)^T&=\frac{1}{z}\left[
            \begin{array}{ccc}
                f_x & 0   & u_0 \\
                0   & f_y & v_0 \\
            \end{array}
        \right]\cdot         
        \left[ \begin{array}{c}x \\
                       y
                       \\
                       z\end{array}\right] \\
        & \bigg|                \; |x-x_{gt}|+|z-z_{gt}| \le \epsilon \Bigg\}.
    \end{aligned}
    \label{eq:prob}
\end{equation}
The above problem is described in \cref{eq:hpe}. $(u_{gt},v_{gt})$ is the ground truth position of an object in the image frame, while $(x_{gt},y_{gt},z_{gt})$ is the ground truth position of the object in the BEV frame.  $S_\epsilon$ is the sampling location set corresponding to the $\epsilon$-neighbourhood of $(x_{gt},z_{gt})$. $y$ is the predicted height at $(x,z)$. Expanding \cref{eq:prob}, we get two equations and one inequality:
\begin{align}
    u_{gt}z=    & f_x x+u_0 z,                      \\
    v_{gt}z=    & f_y y+v_0 z,                      \\
    |x-x_{gt}|+ & |z-z_{gt}|\le\epsilon,\label{ieq}
\end{align}
where $y$ is an abbreviation for $y(x,z)$. According to the transformation between the image frame and the BEV frame:
\begin{align}
    u_{gt}z_{gt}=f_x x_{gt}+u_0 z_{gt}, \\
    v_{gt}z_{gt}=f_y y_{gt}+v_0 z_{gt},
\end{align} we can conclude the relationships among $x-x_{gt}$, $y-y_{gt}$, and $z-z_{gt}$. They follow:
\begin{align}
    (u_{gt}-u_0)(z-z_{gt})=f_x(x-x_{gt}),\label{e1} \\
    (v_{gt}-v_0)(z-z_{gt})=f_y(y-y_{gt}).\label{e2}
\end{align}
Substituting \cref{e1} and \cref{e2} into \cref{ieq}, we can get an inequality about $y$:
\begin{equation}
    \frac{f_y}{f_x}\cdot\frac{|u_{gt}-u_0|}{|v_{gt}-v_0|}|y-y_{gt}|+\frac{f_y}{|v_{gt}-v_0|}|y-y_{gt}| \le \epsilon.
    \label{e3}
\end{equation}
Substituting $y=y_{gt}\pm \delta_y$ into \cref{e3}, we can get the inequality about the height prediction error:
\begin{equation}
    \delta_y\le\epsilon\cdot\frac{|v_{gt}-v_0|}{f_y}\cdot \frac{f_x}{|u_{gt}-u_0|+f_x}.
\end{equation}
As a result, we get the upper bound of the height prediction error:
\begin{equation}
    \delta_{y,max}=\epsilon\cdot\frac{|v_{gt}-v_0|}{f_y}\cdot \frac{f_x}{|u_{gt}-u_0|+f_x},
\end{equation}
which is described in \cref{eq:heighterror}.

}

\section*{Comparison to BEVHeight}

In this appendix, we provide a detailed comparison between the proposed method and BEVHeight\cite{yang2023bevheight}, which also constructs BEV features via height modeling.

\begin{figure}[!h]
    \subfloat[BEVHeight]{
        \includegraphics[width=.8\linewidth,trim=0 100 250 0, clip]{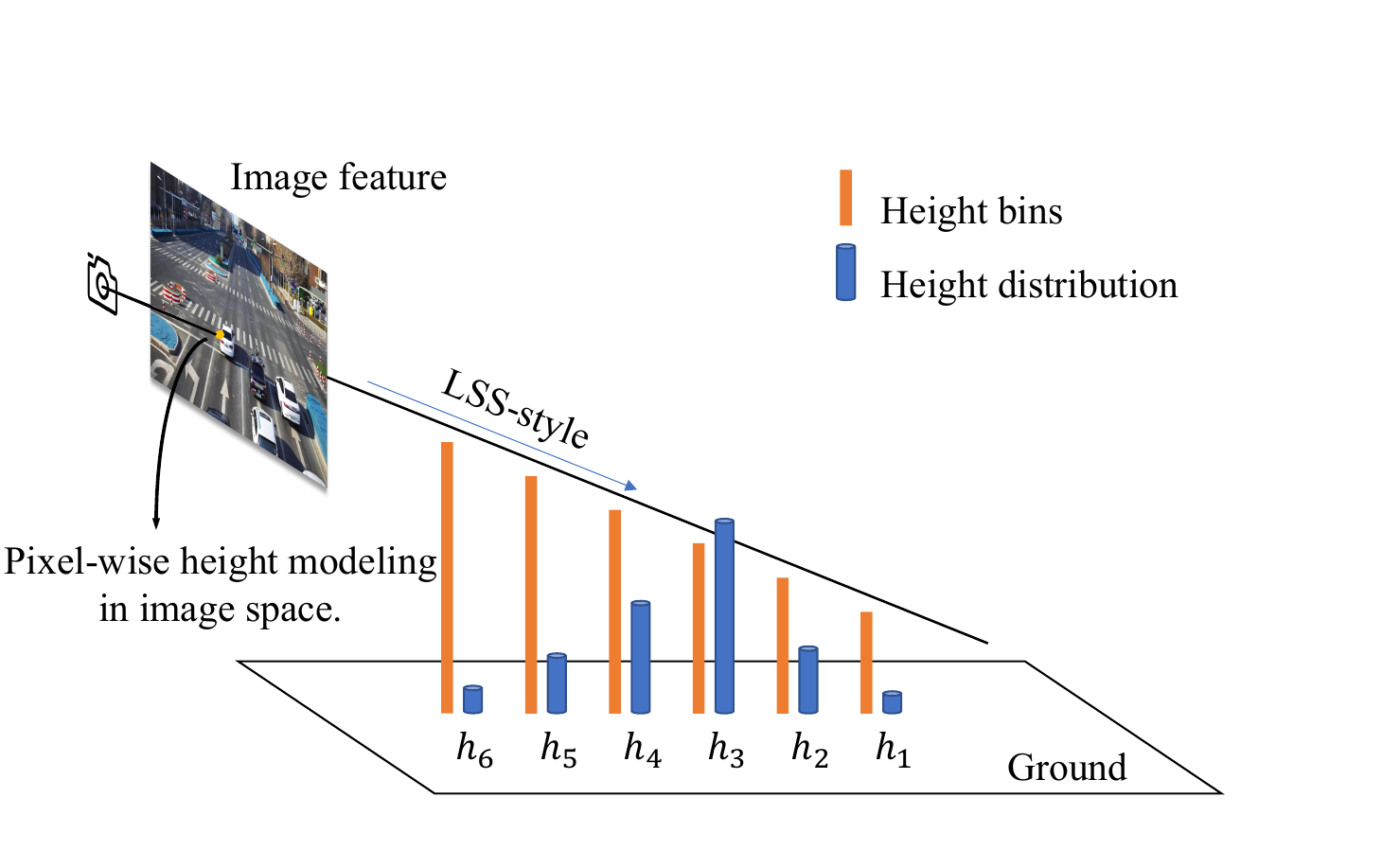}}

    \subfloat[HeightFormer]{
        \includegraphics[width=.8\linewidth,trim=0 100 250 0, clip]{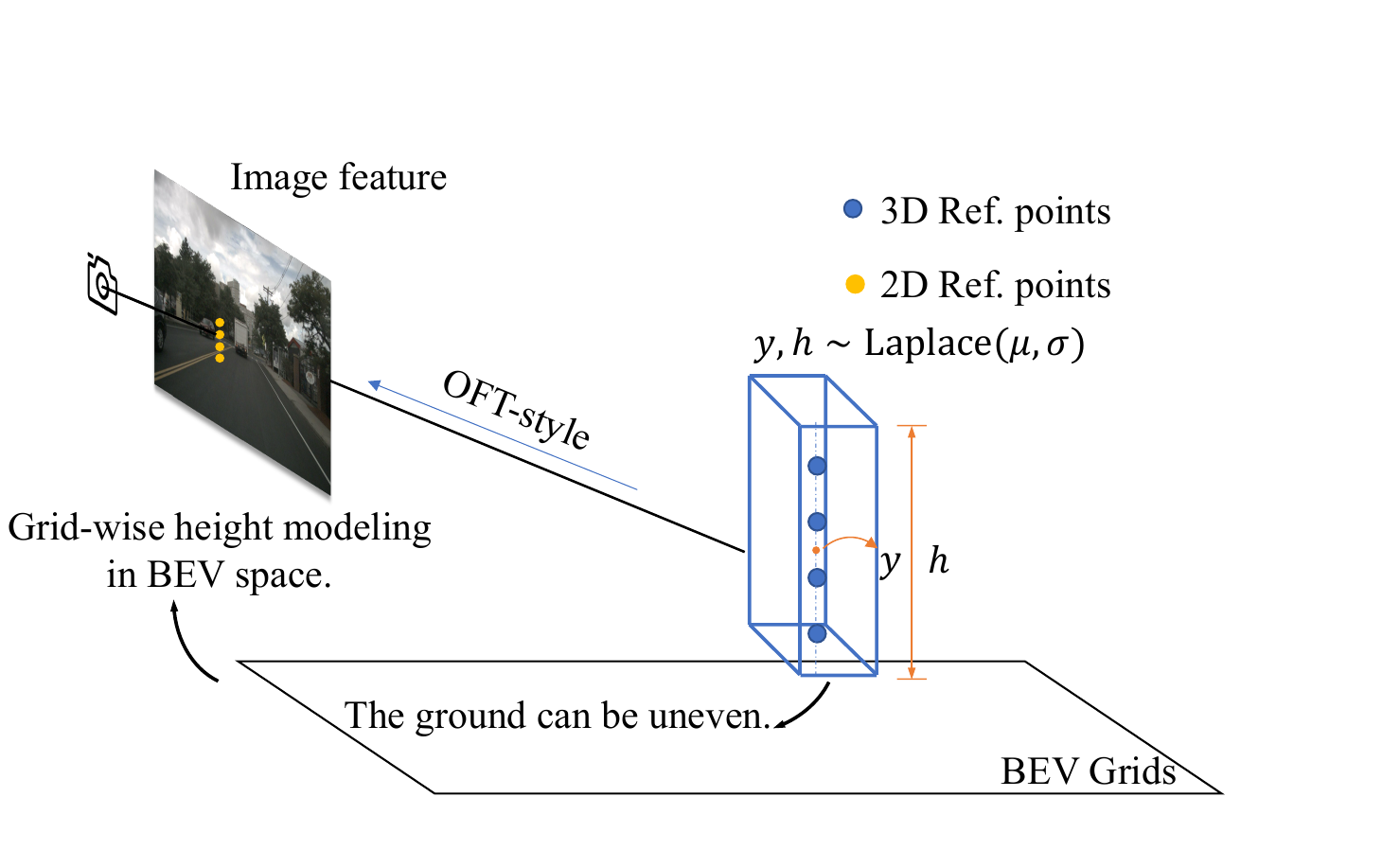}}

    \caption{Differences between BEVHeight and HeightFormer}
    \label{fig1}
\end{figure}

\begin{enumerate}
    \item {\bf Modeling:} BEVHeight models s for each image pixel, and then projects image features into the BEV space. This is {\bf LSS-style}\cite{philion2020lift} modeling, and it can be turned into depth modeling by mapping height bins into depth bins. In contrast, HeightFormer adopts {\bf OFT-style}\cite{roddick2018orthographic} modeling. BEV features are gathered from image features by projecting reference points into images.
    
    \item {\bf Scenario:} BEVHeight is primarily designed for roadside 3D object detection tasks, whereas HeightFormer is tailored for vehicle-mounted scenarios. BEVHeight operates optimally with {cameras with high installation}, as highlighted in its paper. The proposed method has no such requirements as it models heights with BEV features which have historical information about objects.

    \item {\bf Height:} BEVHeight focuses on modeling {\bf surface heights} relative to the { flat ground}, while HeightFormer characterizes both {\bf the center and the range} of an object along the height axis.

    \item {\bf Implementation:} BEVHeight employs a strategy of dividing heights into {\bf discrete} bins and performing classifications for each image pixel. In contrast, HeightFormer utilizes Laplacian priors to model heights in {\bf contiguous} space and conducts height regression for each BEV grid.
\end{enumerate}

\begin{table}[h]
    \centering
    \caption{Comparative experiments with BEVHeight. $\dag$: Official results, single frame model. $*$: Explicit height modeling without other designs or tricks.}
    \begin{tabular}{ccc}
    \toprule
     Model    &   NDS & mAP  \\
     \midrule
      BEVFormer & 0.403 & 0.288  \\
      BEVHeight$^\dag$ & 0.342 & 0.291  \\
      HeightFormer$^*$  & 0.421 & 0.299 \\
      \bottomrule
    \end{tabular}
    \label{tab:vs}
\end{table}

Furthermore, we compare the performance. The \cref{tab:vs} shows the comparative experiments with BEVHeight. Because BEVHeight does not use history frames and its code does not contain training configurations on NuScenes, we compare mAP only. The proposed method outperformed BEVHeight by 0.8 percentage points.


\bibliographystyle{IEEEtran}
\bibliography{IEEEabrv,main}

\end{document}